\documentclass[lettersize,journal]{IEEEtran}
\usepackage{amsmath,amsfonts}
\usepackage{algorithmic}
\usepackage{array}
\usepackage[caption=false,font=normalsize,labelfont=sf,textfont=sf]{subfig}
\usepackage{textcomp}
\usepackage{stfloats}
\usepackage{url}
\usepackage{verbatim}
\usepackage{graphicx}
\def\BibTeX{{\rm B\kern-.05em{\sc i\kern-.025em b}\kern-.08em
    T\kern-.1667em\lower.7ex\hbox{E}\kern-.125emX}}
\usepackage{balance}

\newcommand{\etal}{\textit{et al}.}
\newcommand{\ie}{\textit{i}.\textit{e}.}
\newcommand{\etc}{\textit{etc}}

\usepackage[table]{xcolor} 
\usepackage{times}
\usepackage{epsfig}
\usepackage{graphicx}
\usepackage{amsmath}
\usepackage{amssymb}
\usepackage{booktabs}
\usepackage{bbding}
\usepackage{multirow}
\usepackage{arydshln}

\definecolor{hollywoodcerise}{rgb}{0.96, 0.0, 0.63}
\definecolor{lasallegreen}{rgb}{0.03, 0.47, 0.19}
\definecolor{hanpurple}{rgb}{0.32, 0.09, 0.98}
\definecolor{green(pigment)}{rgb}{0.0, 0.65, 0.31}
\usepackage[pagebackref=true,breaklinks=true,letterpaper=true,colorlinks,bookmarks=false]{hyperref}
\hypersetup{colorlinks,linkcolor={red},citecolor={hanpurple},urlcolor={red}}  
\begin{document}
\title{EventDance++: Language-guided Unsupervised Source-free Cross-modal Adaptation for Event-based Object Recognition}

\author{Xu Zheng,~\IEEEmembership{Student Member,~IEEE,}
        Lin Wang$^{*}$,~\IEEEmembership{Member,~IEEE,}\thanks{* Corresponding Author}
\thanks{Xu Zheng is with the AI Thrust, HKUST(GZ), Guangdong, China. E-mail: zhengxu128@gmail.com.
Lin Wang is with AI/CMA Thrust, HKUST(GZ) and Dept. of CSE, HKUST, Hong Kong SAR, China, E-mail: linwang@ust.hk.
}
}

\markboth{Journal of \LaTeX\ Class Files,~Vol.~18, No.~9, September~2020}%
{How to Use the IEEEtran \LaTeX \ Templates}

\maketitle

\begin{abstract}
In this paper, we address the challenging problem of cross-modal (image-to-events) adaptation for event-based recognition without accessing any labeled source image data. This task is arduous due to the substantial modality gap between images and events. With only a pre-trained source model available, the key challenge lies in extracting knowledge from this model and effectively transferring knowledge to the event-based domain. Inspired by the natural ability of language to convey semantics across different modalities, we propose EventDance++, a novel framework that tackles this unsupervised source-free cross-modal adaptation problem from a language-guided perspective.
We introduce a language-guided reconstruction-based modality bridging (L-RMB) module, which reconstructs intensity frames from events in a self-supervised manner. 
Importantly, it leverages a vision-language model to provide further supervision, enriching the surrogate images and enhancing modality bridging. 
This enables the creation of surrogate images to extract knowledge (\ie, labels) from the source model. On top, we propose a multi-representation knowledge adaptation (MKA) module to transfer knowledge to target models, utilizing multiple event representations to capture the spatiotemporal characteristics of events fully. The L-RMB and MKA modules are jointly optimized to achieve optimal performance in bridging the modality gap.
Experiments on three benchmark datasets demonstrate that EventDance++ performs on par with methods that utilize source data, validating the effectiveness of our language-guided approach in event-based recognition.
Project Page: \url{https://vlislab22.github.io/EventDanceplus/}.
\end{abstract}


\begin{figure}[t!]
    \centering
    \includegraphics[width=\linewidth]{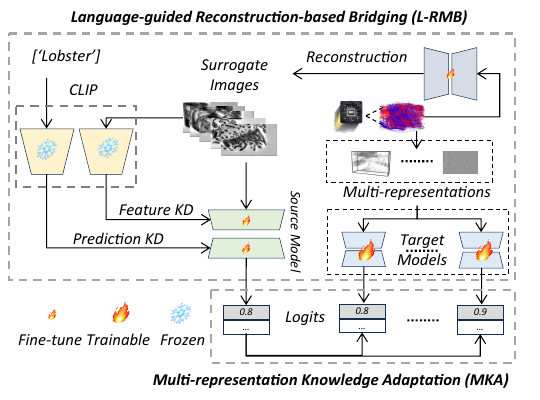}
    \vspace{-12pt}
    \caption{Illustration of the challenging task of cross-modal adaptation from image to event modalities.
    We address it by introducing language-guided reconstruction-based modality bridging and multi-representation knowledge adaptation modules.}
    \label{fig:Teaser_figure}
    \vspace{-12pt}
\end{figure}

\section{Introduction}
\IEEEPARstart{E}{vent} cameras, also known as silicon retina~\cite{retinomorphic}, are bio-inspired sensors that asynchronously detect per-pixel intensity changes, producing streams of events that encode the time, pixel position, and polarity of these changes~\cite{eventsurvey, EST, Hots}. Event cameras offer several advantages, such as high dynamic range and low latency, making them particularly useful in challenging visual conditions where conventional frame-based cameras fall short~\cite{deeplearningeventsurvey, RealtimeHSpeed, TORE, SITS, SpeedHDR, SLAMHSpeed, speedHDRrecon}. Consequently, these sensors have attracted considerable attention in computer vision and robotics research.

However, the development of deep learning models for event cameras has been limited by the scarcity of large-scale, labeled datasets due to the asynchronous and sparse nature of event data. This makes it difficult to apply traditional supervised learning approaches. As a result, recent research has explored on cross-modal adaptation strategies to transfer knowledge from labeled image data (\ie, source) to unlabeled event data (\ie, target)~\cite{CTN, wang2021evdistill, messikommer2022bridging}.
At the same time, in real-world applications, access to source datasets is often restricted due to privacy concerns and commercial barriers, such as data protection regulations and high portability costs~\cite{SFUDAsurvey}.

In this paper, we address a novel challenge of cross-modal (\ie, image-to-events) adaptation for event-based recognition \textit{without access to any labeled source image data}. 
In this setting, we aim to leverage a pre-trained source model (trained on images) and adapt it to the event-based recognition. 
The significant modality gap between images and events complicates this task, especially since only the source model is available. The key problem is: \textit{\textbf{how can we achieve efficient knowledge extraction from the source model and then transfer the extracted knowledge to the event-based vision with only event data?}}


Inspired by the fact that language naturally conveys semantic information across modalities, we propose \textbf{EventDance++}, a novel framework that, for the first time, approaches this unsupervised source-free cross-modal adaptation problem from a language-guided perspective.
Bridging the substantial modality gap with language poses technical challenges, particularly in aligning semantic representations with the sparse and asynchronous nature of event data.
To this end, we define two primary learning objectives: \textbf{(1)} bridging the modality gap between images and events, and \textbf{(2)} transferring knowledge from the source model to the target event domain. To this end, we introduce a language-augmented reconstruction-based modality bridging (\textbf{L-RMB}) module (Sec.~\ref{sec: Modality Bridging}), which reconstructs intensity frames from events in a self-supervised manner, enabling the generation of surrogate images for knowledge extraction (\ie, labels) from the source model. This reconstruction step effectively reduces the large modality gap between the image and event domains.

Unlike traditional reconstruction approaches that prioritize generating visually realistic images~\cite{BackEventBasic}, our L-RMB module is specifically fine-tuned to enhance knowledge extraction from the source model by optimizing for lower entropy in predictions and ensuring temporal consistency, leveraging the high temporal resolution of event data. Additionally, we incorporate a vision-language model (\ie, CLIP) to provide additional supervision, using language-guided signals to improve the quality of surrogate images and facilitate more efficient knowledge transfer. By introducing feature-level and prediction-level knowledge distillation constraints, we further enhance knowledge extraction from the source model.

We then propose a Multi-Representation Knowledge Adaptation (MKA) module (Sec.~\ref{sec: Knowledge Adaptation}) that facilitates knowledge transfer to target models learning from unlabeled events. As substantial information loss, such as timestamp drops, occurs when converting events to specific representations like event count images, it may impede object recognition performance~\cite{deeplearningeventsurvey}. Consequently, we employ multiple event representations in our EventDance++ framework, including stack images~\cite{stackimage}, voxel grids~\cite{voxelgrid}, and event spike tensors (EST)~\cite{EST}, to fully harness the spatiotemporal information in events. This strategy ensures cross-modal prediction consistency during target model training. The two modules linking the source and target models are iteratively updated to optimize modality bridging and knowledge adaptation.

We validate EventDance++ on three prominent event-based recognition benchmarks: N-Caltech101~\cite{dataset}, N-MNIST~\cite{dataset}, and CIFAR10-DVS~\cite{li2017cifar10}. 
The experimental results show that EventDance++ significantly surpasses existing source-free domain adaptation methods, such as~\cite{SHOT}, in tackling this challenging cross-modal task.

In summary, our key contributions are: \textbf{(I)} We tackle a novel and challenging cross-modal (image-to-events) adaptation problem without access to any source image data. \textbf{(II)} We propose EventDance++, a framework that leverages the L-RMB and MKA modules to bridge the modality gap and transfer knowledge effectively between images and events. \textbf{(III)} We introduce a vision-language model to guide the knowledge extraction process through language-augmented supervision. \textbf{(IV)} Extensive experiments on three event-based benchmarks demonstrate the superiority of EventDance++ over existing methods in this challenging domain.

\begin{figure}[t!]
    \centering
    \includegraphics[width=\linewidth]{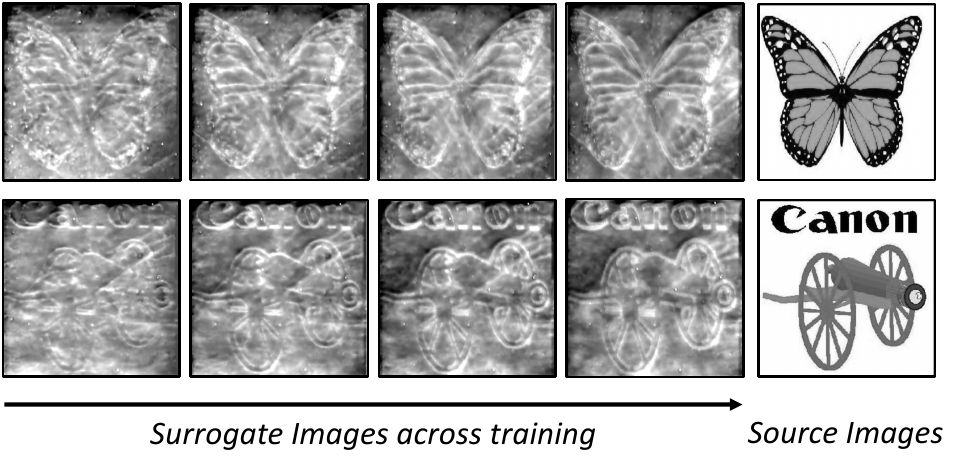}
    \vspace{-12pt}
    \caption{The reconstructed surrogate images across the knowledge adaptation. 
    }
    \label{fig:loss_LR}
    \vspace{-12pt}
\end{figure}

This work extends our CVPR 2024 paper~\cite{zheng2024eventdance}, incorporating technical and experimental updates in the following aspects: 
\begin{itemize} \item We introduce the Language-guided Reconstruction-based Modality Bridging (L-RMB) module (Sec.~\ref{sec: Modality Bridging}), which self-supervises the reconstruction of surrogate intensity frames from events. By leveraging a vision-language model, we enrich the surrogate images with additional supervision, significantly enhancing the modality bridging between events and images. 
\item The knowledge distillation is performed in both feature- and prediction-level from the text and image encoders of the CLIP model with corresponding loss constraints to improve source knowledge extraction (Sec.~\ref{sec: Modality Bridging} 3)). 
\item We reformulate the overall loss functions to simplify the training process and eliminate the need for extensive hyper-parameter tuning, optimizing the knowledge transfer between modalities. 
\item We conduct additional comparative experiments on three cross-modal benchmarks (Tab.~\ref{tab:MNIST-to-N-MNIST},~\ref{tab:Caltech101-to-N-Caltech101}, and~\ref{tab:CIFAR10-to-N-CIFAR10},), thoroughly evaluating the new framework against both the previous version and state-of-the-art SFUDA methods. \item Extensive quantitative and qualitative analyses, as seen in Fig.~\ref{fig:tsne_caltech},~\ref{fig:tsne_compare_cifar}, and~\ref{fig:tsne}, and~\ref{fig:loss_LR}, validate the effectiveness of the newly introduced strategies and components. \end{itemize}


\begin{figure}[t!]
    \centering
    \includegraphics[width=0.35\textwidth]{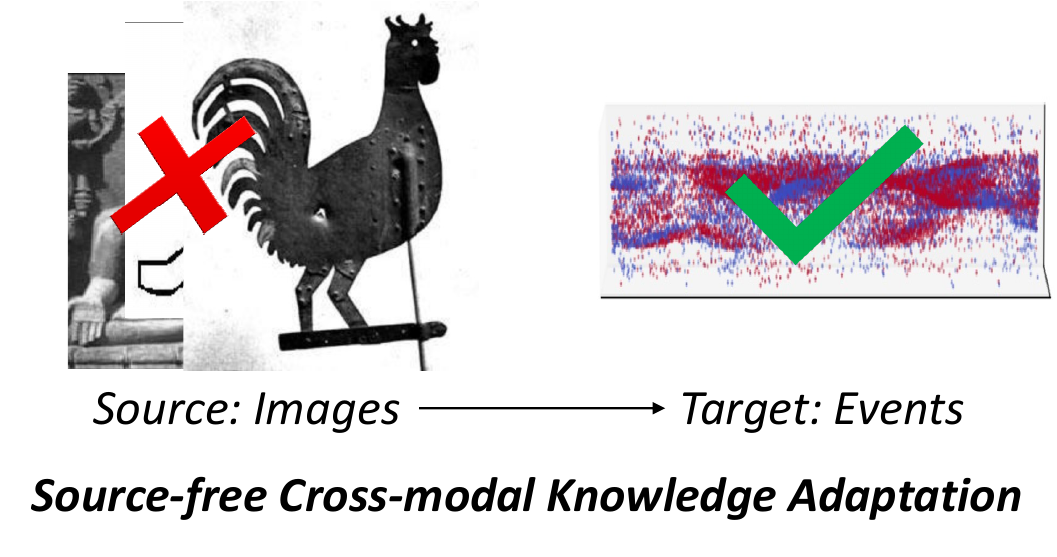}
    \vspace{-12pt}
    \caption{\textbf{Cross-modal knowledge adaptation settings}. 
   }
    \vspace{-12pt}
    \label{fig:exp_set_com} 
\end{figure}

\begin{figure*}[h!]
    \centering
    \includegraphics[width=0.88\textwidth]{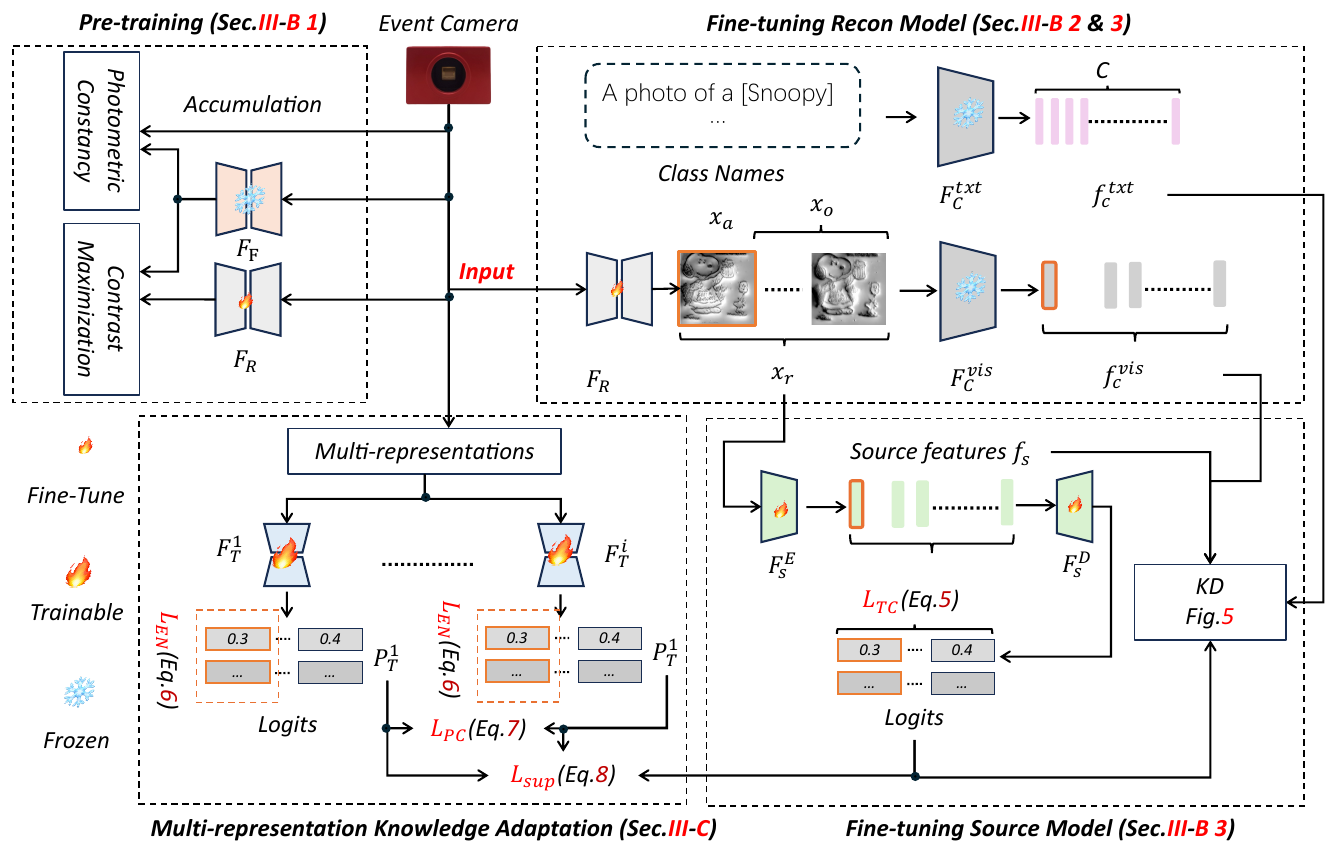}
    \vspace{-4pt}
    \caption{Overall framework of our proposed SFUDA for panoramic semantic segmentation. 
    }
    \label{fig:overall}
    \vspace{-8pt}
\end{figure*}
\section{Related Work}
\subsection{Event-based Object Recognition}
Event-based object recognition seeks to identify target objects from event streams, capitalizing on the unique attributes of event cameras. Owing to their high temporal resolution, low latency, and extensive dynamic range, event cameras facilitate real-time onboard object recognition in applications such as robotics, autonomous vehicles, and other mobile systems~\cite{deeplearningeventsurvey}. However, the distinctive imaging paradigm of event cameras precludes the direct application of deep neural networks (DNNs) for event learning. Consequently, a variety of event representations have been developed~\cite{maqueda2018event, EvFlowNet, wang2019ev, deng2020amae, deng2021mvf, gehrig2019end, cannici2020differentiable, almatrafi2020distance, deng2022voxel}, aimed at effectively extracting visual information from events, particularly for object recognition tasks. Previous research, such as~\cite{CTN}, has utilized diverse event representations as the target domain, yet these studies have not fully leveraged the potential of raw events. \textit{In this work, we propose learning target models that not only differentiate raw events but also employ multiple event representations to enforce consistency regularization.}

Recently, the CLIP model has been adapted for use in event-based vision, particularly within the object recognition domain~\cite{zhou2024eventbind,cho2023label}. EventBind~\cite{zhou2024eventbind} represents a notable attempt to develop an event encoder specifically for the event modality. In contrast, our approach with EventDance++ fundamentally diverges in how we utilize the CLIP model. We employ the CLIP model as a teacher for the source model, implementing a knowledge distillation strategy to enhance the source model’s capabilities. This approach facilitates more effective knowledge transfer between modalities.

\subsection{Cross-modal Knowledge Transfer}
Knowledge transfer across modalities is first proposed in ~\cite{CMKD}, with the objective of learning representations for modalities constrained by limited annotations by leveraging data from a label-rich modality. There has been a surge of interest in applying cross-modal knowledge transfer to innovative sensor technologies, such as event cameras~\cite{CMKDAR, CMKDAD, CMARD}. While most established methods presuppose the availability of cross-modal paired data, recent studies have attempted to mitigate this requirement by minimizing the necessary data volume~\cite{KAP}. Moreover, several approaches have explored classification through domain translation strategies~\cite{DTRGBD, T2ARGBD}, further expanding the scope of cross-modal knowledge transfer applications.
These methods depend on cross-modal data pairs and task-relevant paired data. To reduce the dependency on such paired data, SOCKET~\cite{SOCKET} introduces a cross-modal adaptation framework that leverages only externally sourced extra paired data for RGB-to-depth knowledge transfer, which can be challenging to procure.
\textit{Differently, our EventDance++ represents the first framework designed for cross-modal (image-to-event) adaptation without access to any source modality data. }

\subsection{Source-free Unsupervised Domain Adaptation}
UDA aims to alleviate the domain-shift problems caused by data distribution discrepancy in many computer vision tasks~\cite{CCUDA, LUDA, UDA-COPE, DUDA, DAWL, UMTDA, UDANAT, UDACAS, SUDA, zheng2023both, zheng2023look,zheng2024semantics,zheng2024360sfuda++,zhang2024goodsam++}. However, the dependence on source data limits the generalization capability to some real applications, for reasons like data privacy issues~\cite{SHOT}. Thus endeavors have been made in transferring knowledge only from the trained source models~\cite{HTL} without access to the source data. The cross-domain knowledge for unlabeled target data is extracted from single~\cite{LPSUDA} or multiple~\cite{MSUDA} source models without access to the source data~\cite{SHOT}. The ideas of source-free UDA can be formulated into two types according to whether the parameters of source models are available~\cite{SFUDAsurvey}, \ie, white-box and black-box models. 
Concretely, the white-box are achieved by data generation~\cite{SFUDAsurrogate, SFUDAgenerate, SFDAestimation, UDAcontinual} and model fine-tuning~\cite {SFUDAdiversification, SFUDAbatchnorm} while the black-box depend on self-supervised learning~\cite{DINE, MutialNet} and distribution alignment~\cite{ITRL, DJP-MMD}. 

CTN~\cite{CTN} is a UDA framework that leverages the edge maps obtained from the source RGB images and adapts the classification knowledge to a target model learning event images.
In this paper, we focus on the source-free cross-modal (\ie, image-to-event) adaptation without accessing the source data, which is essentially different from~\cite{CTN} and more challenging to tackle. \textit{Our core idea is to create a surrogate domain in the image modality via the L-RMB module and update the surrogate domain for knowledge transfer via the MKA module.}

\section{The Proposed Framework}
We first describe the our problem setting in Sec.\ref{sec:Problem Setup}. 
Then, we elaborate the framework of our method (see Fig.~\ref{fig:overall}): 
Language-guided Reconstruction-based Modality Bridging (L-RMB) (Sec.~\ref{sec: Modality Bridging}) and Multi-representation Knowledge Adaptation (MKA) (Sec.~\ref{sec: Knowledge Adaptation}).

\subsection{Problem Setup and Overview}
\label{sec:Problem Setup}
Transferring knowledge from a source modality to a target modality always presents more significant challenges than managing domain shifts within the same modality across different datasets. This increased complexity has been underscored by findings presented in \cite{SOCKET}. Traditional approaches to cross-modal adaptation, as explored in \cite{DTRGBD, T2ARGBD, SOCKET}, generally rely on supplementary data from both source and target modalities to facilitate the adaptation process. However, the distinct characteristics of event cameras, coupled with the lack of available paired data, significantly limit the applicability of these conventional methods to the event modality. Consequently, within the context of our cross-modal problem, we are constrained to utilize solely a pre-trained source model from the image modality and unlabeled target data from the event modality. In contrast to previous research such as \cite{SOCKET}, our study does not utilize any external datasets. Our experimental design is confined to using only a pre-trained source model specific to the image modality and unlabeled target data within the event modality. 

\noindent \textbf{Our Key Idea}: \textit{By constructing surrogate domain with target events, we aim to mitigate modality gaps with the guidance from VLM. This enables better knowledge extraction from the source model. We subsequently employ multiple representations of the events and harness language as guidance to accomplish the knowledge transfer.}

\subsubsection{Primary Objective} 
\begin{figure}[t!]
    \centering
    \includegraphics[width=\linewidth]{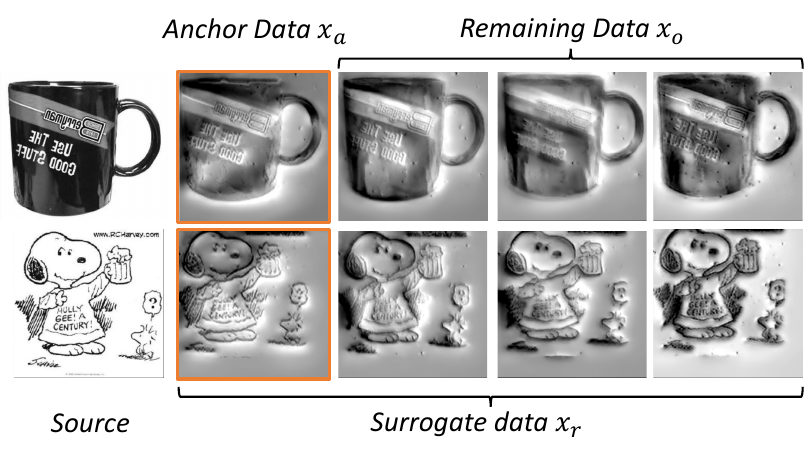}
    \vspace{-20pt}
    \caption{(a) Example visualization of samples in the source (gray-scale image) and the surrogate (reconstructed) data in the image modality. (b) The reconstructed anchor data from the surrogate data across the knowledge adaptation.}
    \vspace{-12pt}
    \label{fig:Recon_com}
\end{figure}

Denote the source model as \( F_S \), where \( S \) indicates the source modality on which the model is trained. Let \( X_T \) represent the unlabeled target event data, where \( T \) denotes the target modality. As depicted in Fig.~\ref{fig:overall}, given a batch of event data \( x_t \subset X_T \), we derive the surrogate image batch \( x_r \) using a self-supervised, pre-trained reconstruction model \( F_R \). An example of the reconstructed surrogate data is illustrated in Fig.~\ref{fig:Recon_com}. Our objective is to infer target models \( F_T^i \) using the source model \( F_S \) and the unlabeled event data \( \mathbf{X_T} \), while distilling knowledge from the CLIP text and visual encoders \( F_C^{\text{txt}} \) and \( F_C^{\text{vis}} \).
Here, the index \( i \) indicates the \( i \)-th target model, which ingests different forms of event representation as input. Specifically, for \( i=1 \), the input is a stack image; for \( i=2 \), it is a voxel grid; and for \( i=3 \), it is an event spike tensor (EST). 
In subsequent sections, we elaborate on the proposed modules: the Language-guided Reconstruction-Based Modality Bridging (L-RMB) module (Sec.~\ref{sec: Modality Bridging}) and the Multi-Representation Knowledge Adaptation (MKA) module (Sec.~\ref{sec: Knowledge Adaptation}).

\subsection{Language-guided Reconstruction-based Modality Bridging (L-RMB)}
\label{sec: Modality Bridging}

\subsubsection{Self-supervised Pre-training}
The L-RMB module creates a surrogate image domain that closely mirrors the distribution of the source image modality. Utilizing a self-supervised event-to-video model \cite{BackEventBasic}, this module generates surrogate data directly from raw event data, as illustrated in Fig.~\ref{fig:Recon_com}. This surrogate data is pivotal for extracting knowledge, specifically pseudo-labels, from the source model that has been trained on image data from the source modality.
\textit{However, the mere utilization of such a model does not fully satisfy our objectives}. A primary drawback of this approach is its emphasis on producing natural-looking images, rather than optimizing the generation of surrogate images that are better suited for facilitating effective knowledge extraction from the source model. Consequently, we propose to fine-tune the L-RMB module during the training process to improve the quality of surrogate image generation, thereby optimizing them for more effective knowledge extraction.

In practice, we employ the framework outlined in \cite{BackEventBasic} as our foundational framework, which incorporates EvFlowNet \cite{EvFlowNet} as the flow estimation model \( F_F \) and E2VID \cite{E2VID} as the reconstruction model \( F_R \). EvFlowNet is trained utilizing a contrast maximization proxy loss \cite{Contrast}, enabling it to provide accurate optical flow estimates. E2VID, on the other hand, reconstructs intensity frames by exploiting the relationship between flow and intensity, adhering to the principles of event-based photo-metric constancy \cite{Photometric}.
Both \( F_F \) and \( F_R \) are initially pre-trained using unlabeled event data. During the our EventDance++ training process, only \( F_R \) is updated, guided by the loss functions \( \mathcal{L}_{EN} \).

\subsubsection{CLIP Feature Extraction}
Recent studies, such as \cite{zhou2024exact,zhou2023clip}, have highlighted the significant advantages of the CLIP model for event-based vision tasks. Vision-Language Models (VLMs) like CLIP are essential in this domain. They bridge the gap between visual and textual modalities, enabling richer and more comprehensive feature extraction.
Incorporating language guidance allows the model to generalize better across modalities. \textit{It helps the model interpret the reconstructed images from raw event data by linking them with clearly defined textual categories.} This compensates for the lack of dense visual cues that are typically present in traditional event representations. By leveraging our reconstruction-based modality bridging module alongside CLIP, we enhance the feature space with meaningful semantic information. This facilitates more accurate and efficient cross-modal knowledge transfer between the image and event domains.

\textbf{\textit{Building on our EventDance~\cite{zheng2024eventdance}, we integrate the CLIP model into the enhanced EventDance++ to improve the source model $F_S$ for better source knowledge extraction.}}
As illustrated in Fig.~\ref{fig:overall} \textcolor{red}{(b)}, the process begins with the creation of textual prompts for each class name, formatted as \texttt{"A photo of [Dragonfly]"}, which are then processed by the CLIP text encoder \( F_C^{txt} \) to extract textual features \( f_c^{txt} \in \mathbb{R}^{K \times 768} \), where \( K \) represents the total number of categories. Concurrently, the surrogate images \( x_r \) are fed into the CLIP visual encoder \( F_C^{vis} \), resulting in visual features \( f_c^{vis} \in \mathbb{R}^{n \times 768} \), where \( n \) denotes the number of reconstructed images \( x_r \).

\subsubsection{Fine-tuning Reconstruction \& Source Model}
In the subsequent phase, we employ the reconstruction model \( F_R \) to convert raw events captured by the event camera into gray-scale images \( x_r \). This set includes the initial frame \( x_a \), designated as the anchor data, and subsequent frames \( x_o \). These reconstructed images \( x_r \) are then fed into the source model \( F_S \) to facilitate knowledge extraction. For predictions derived from the anchor data \( F_{S}(x_a) \), we apply entropy minimization to ensure that the surrogate images are optimized for effective knowledge extraction from the source model. The fine-tuning of the reconstruction model \( F_R \) is governed by the loss function \( \mathcal{L}_{R} \), defined as follows:
\begin{equation}
\label{eq:L_R}
    \mathcal{L}_{R} = \min (H(F_S(x_a))),
\end{equation}
where \( H(\cdot) \) denotes the entropy function. In Fig.~\ref{fig:Recon_com} (b), we illustrate the reconstructed images across the knowledge adaptation.

\begin{figure}[t!]
    \centering
    \includegraphics[width=\linewidth]{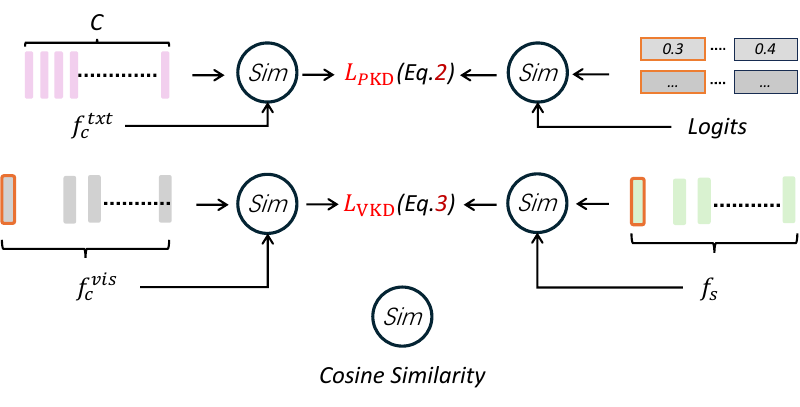}
    \vspace{-16pt}
    \caption{Fine-tuning source model with both feature-wise and prediction-wise knowledge distillation. 
    }
    \label{fig:KD}
    \vspace{-12pt}
\end{figure}

Furthermore, we exploit CLIP features as supervisory signals to augment the source model for better knowledge extraction at both the feature and prediction levels, utilizing the extracted textual and visual features, \( f_c^{txt} \) and \( f_c^{vis} \), respectively. As shown in Fig.~\ref{fig:KD}, Specifically, the visual features \( f_c^{vis} \) are employed to supervise the source features \( f_s \) through our proposed knowledge distillation loss, \( \mathcal{L}_{VKD} \), defined as:
\begin{equation}
\label{eq:L_vkd}
    \mathcal{L}_{VKD} = \text{KL}(f_s \times f_s^T, \text{softmax}(f_c^{vis} \times (f_c^{vis})^T)),
\end{equation}
where \( \text{KL} \) denotes the Kullback-Leibler divergence. It facilitates the transfer of complex spatial correlations captured by the CLIP visual encoder to the source model, thus enabling a more sophisticated and robust representation of features.

Subsequently, the textual features, \( f_c^{txt} \), are employed to provide supervision at the prediction level for the source model using the prediction-level knowledge distillation loss, \( \mathcal{L}_{PKD} \):
\begin{equation}
\label{eq:L_pkd}
    \mathcal{L}_{PKD} = \text{KL}(P_s^a \times (P_s^a)^T, \text{softmax}(f_c^{txt} \times (f_c^{txt})^T)),
\end{equation}
This dual-level supervision strategy ensures that the source model not only better aligns with the semantic attributes encoded by the CLIP model but also enhances the predictive consistency across diverse inputs.
The cumulative knowledge distillation loss is defined as:
\begin{equation}
\label{eq:L_kd}
    \mathcal{L}_{KD} = \mathcal{L}_{PKD} + \mathcal{L}_{VKD},
\end{equation}
As illustrated in Fig.~\ref{fig:Recon_com} (a), we further exploit the remaining data $x_o$ (excluding the selected anchor data $x_a$) to augment the anchor data $x_a$, thereby fully leveraging the high-temporal resolution characteristic of event cameras. To ensure temporal prediction consistency among the reconstructed images, we refine $F_{S}$ using the temporal consistency loss $\mathcal{L}_{TC}$:
\begin{equation}
\label{eq:L_tc}
    \mathcal{L}_{TC} = \mathcal{L}_{kl}(F_{S}(x_a), F_S(x_o)),
\end{equation}
where $\mathcal{L}_{kl}$ denotes the Kullback-Leibler (KL) divergence.
With the knowledge extraction module, we obtain the prediction logits $P$ from $F_S(x_a)$ for learning the target event-based models in the following knowledge adaptation module.

\subsection{Multi-representation Knowledge Adaptation}
\label{sec: Knowledge Adaptation}
Although our L-RMB module is instrumental in bridging modality gaps, the task of adapting knowledge from images to event-based data presents significant challenges. These include: \textbf{1)} the limitation of a single event representation type, \textit{e.g.}, voxel grids~\cite{CTN}, which may not capture the full complexity of event data, potentially leading to substantial information loss during the adaptation process, and \textbf{2)} inefficiencies in the source model that may impair its effectiveness in cross-modal knowledge transfer. To address these issues, we propose a strategy to concurrently learn multiple target models, each utilizing a distinct event representation. This approach is designed to maximize the utilization of the intrinsic high temporal resolution of event data, as depicted in Fig.~\ref{fig:overall}. This multi-representation framework not only enhances the fidelity of event data representation but also improves the robustness and efficiency of knowledge transfer across modalities.

To process raw events, we convert the incoming event stream \(E\) into commonly utilized representations. For the voxel grid representation, following the methodology proposed in \cite{voxelgrid}, we construct the voxel grid \(E_v \subset \mathbb{R}^{H \times W \times C}\), partitioning \(E\) into \(B\) temporal bins. These bins consist of consecutive, non-overlapping segments of the event stream, where \(H\), \(W\), and \(C\) denote the spatial dimensions. The voxel grid \(E_v\) adaptively normalizes the temporal dimension based on the timestamps associated with each segment, thereby preserving temporal information within the spatial framework.
For the representation of event stack images, we implement a predefined stacking strategy as detailed in \cite{stackimage}, where events are sampled and stacked until a fixed count threshold is reached, yielding a tensor-like representation \(E_s \subset \mathbb{R}^{H \times W \times 1}\). This approach ensures that spatial details are effectively maintained while capitalizing on the high temporal resolution characteristic of events.
For the Event Spike Tensor (EST), we employ the method described in \cite{EST}, which processes raw events directly.

For the predictions based on the anchor data \( F_{T}^i(x_a) \), we implement entropy minimization for better target model training. The optimization of the target model \( F_T^i \) is first driven by the entropy minimization loss function \( \mathcal{L}_{EN} \), which is defined as follows:
\begin{equation}
\label{eq:L_en}
    \mathcal{L}_{EN} = \min (H(F_S(x_a))),
\end{equation}
where \( H(\cdot) \) denotes the entropy function. This approach is intended to refine the predictions of \( F_T^i \) by reducing the entropy of the source model's predictions on the surrogate images, thereby enhancing the predictive alignment and the overall effectiveness of knowledge transfer.

Furthermore, as illustrated in Fig.~\ref{fig:overall}, we establish two additional training objectives for the target models: \textbf{1)} \textit{cross-representation consistency} training, which employs various event representation types to enhance the robustness and generalizability of the target models \( F_{T}^i \); and \textbf{2)} \textit{cross-modal consistency} training, aimed at synchronizing the knowledge transfer between the source model \( F_{S} \) and each \( i \)-th target model \( F_{T}^i \). These strategies are designed to ensure that the models not only learn from diverse representations within the same modality but also maintain alignment across different modalities, thereby improving the overall efficacy of the adaptation process.

\noindent \textbf{Cross-representation Consistency.} Three distinct event representation types are fed into their corresponding target models, upon which prediction consistency training is initiated to facilitate mutual learning among the models. This training aims to harmonize the learning trajectories of the models by minimizing the divergence in their predictive outputs. Such synchronization is intended to foster a more robust understanding of the event data. The prediction consistency training loss, \( \mathcal{L}_{PC} \), which measures the discrepancy across these different event representations, is mathematically expressed as follows:
\begin{equation}
\label{eq:prediction_consistency_loss}
    \mathcal{L}_{PC} = \sum_{k=1}^{3} \sum_{\substack{l=1 \\ l \neq k}}^{3} \left\{\mathcal{L}_{kl}(F_T^{k}(r(x_{t})^{k}), F_T^{l}(r(x_{t})^{l}))\right\}.
\end{equation}

\noindent \textbf{Cross-modal Consistency.} 
To facilitate the transfer of knowledge from the source modality to the target modality, we employ the prediction logits \( P \) obtained from the source model as a supervisory signal for training the target models. We denote the \( i \)-th event representation input to the target model \( F_{T}^i \) as \( r(x_{t})^i \). The supervision process of the target models by the prediction logits \( P \) is formalized as follows:
\begin{equation}
\label{eq:L_sup}
    \mathcal{L}_{Sup} = KL(F_{S}(r(x_{t})^i), \text{softmax}(P)), \quad i \in \{1,2,3\}.
\end{equation}
Here, \( \mathcal{L}_{Sup} \) represents the cross-modal knowledge transfer loss, which employs the Kullback-Leibler divergence to synchronize the outputs of the target models with the predictions from the source modality, thereby enabling effective knowledge adaptation across modalities.

\noindent Overall, our final loss function, \( \mathcal{L}_{all} \), is formulated by aggregating the various loss components specified in Eq.~\ref{eq:L_en}, ~\ref{eq:L_vkd}, ~\ref{eq:L_tc}, and ~\ref{eq:L_sup}. The overall loss function is defined as:
\begin{equation}
    \mathcal{L}_{all} =  \mathcal{L}_{R} + \mathcal{L}_{KD} + \mathcal{L}_{TC} + \mathcal{L}_{EN} + \mathcal{L}_{PC} + \mathcal{L}_{Sup}.
\end{equation}
Specifically, \( \mathcal{L}_{R} \) is utilized to optimize \( F_R \); \( \mathcal{L}_{KD} \) and \( \mathcal{L}_{TC} \) target the optimization of \( F_{S} \); and \( \mathcal{L}_{EN} \), \( \mathcal{L}_{PC} \), and \( \mathcal{L}_{Sup} \) are employed to refine \( F_{T}^i \). The entire framework is optimized in an end-to-end manner, facilitating seamless integration and synchronization of the various components.

\begin{table*}[ht!]
	\centering
	\caption{Experimental results on images-to-events (MNIST-to-N-MNIST) with \textbf{SFUDA} methods. 
		$\Delta$: The performance gain over the baseline. 
		The \textbf{bold} and \underline{underline} denote the best and the second-best performance in SFUDA methods, respectively.}
	\renewcommand{\tabcolsep}{12pt}
	\resizebox{\linewidth}{!}{
		\begin{tabular}{lcclcccl}
			\toprule
			Method & Backbone & Accuracy & $\Delta$ & Recall & $\Delta$ & F1 & $\Delta$ \\ \midrule
			Baseline & & 41.03 & - & 27.05 & - & 27.03 & -  \\ 
			SHOT ~\cite{SHOT} & & 52.60 & +11.57 & 34.49 & +7.44 & 32.19 & +5.16 \\ 
			Zhao \etal ~\cite{CTN} & & 54.08 & +13.05 & 37.37 & +10.32 & 34.94 & +7.91 \\
			SHOT++ ~\cite{SHOT++} & ResNet-18 & 68.87 & +27.84 & 45.50 & +18.45 & 39.56 & +12.53 \\ 
			EventDance~\cite{zheng2024eventdance} & & 70.35 & +29.32 & 46.09 & +19.04 & 40.55 & +13.52 \\ 
			\rowcolor{gray!10} EventDance++ (Ours) & & 75.33 & +34.30 & 51.72 & +24.67 & 54.14 & +27.11 \\ 
			\textit{w.r.t.} EventDance~\cite{zheng2024eventdance} & & & \textbf{+4.98 }& & \textbf{+5.63} & & \textbf{+13.59} \\
			\midrule
			Baseline & & 65.86 & - & 59.84 & - & 56.71 & - \\ 
			SHOT ~\cite{SHOT} & & 75.20 & +9.34 & 69.42 & +9.58 & 67.42 & +10.71 \\ 
			Zhao \etal ~\cite{CTN}& & 76.61 & +10.75 &  65.25& +5.41 & 64.71 & +8.00 \\ 
			SHOT++ ~\cite{SHOT++} & EfficientNet & 77.91 & +12.05 & 72.34 & +12.50 & 70.40 & +13.69  \\ 
			EventDance~\cite{zheng2024eventdance} & & 84.94 & +19.08 & 83.46 & +23.62 & 82.98 & +26.27 \\ 
			\rowcolor{gray!15} EventDance++ (Ours)  & & 85.16 & +19.30 & 83.73 & +23.89 & 83.26 & +26.55 \\ 
			\textit{w.r.t.} EventDance~\cite{zheng2024eventdance} & & & \textbf{+0.22} & & \textbf{+0.27} & & \textbf{+0.28} \\
			\midrule
			Baseline & & 57.36 & - & 42.69 & - & 34.95 & -  \\ 
			SHOT ~\cite{SHOT} & & 68.46 & +11.10 & 45.03 & +2.34 & 36.18 & +1.23 \\ 
			Zhao \etal ~\cite{CTN}&  & 68.02 & +10.66 & 43.84 & +1.15 & 34.56 & -0.39  \\ 
			SHOT++ ~\cite{SHOT++} & ConvNext-tiny & 68.46 & +11.10 & 45.12 & +2.43 & 33.97 & -0.98 \\ 
			EventDance~\cite{zheng2024eventdance} & & 70.36 & +13.00 & 51.98 & +9.29 & 44.12 & +9.17   \\ 
			\rowcolor{gray!15} EventDance++ (Ours)  & & 71.31 & +13.95 & 52.05 & +9.36 & 44.32 & +9.37 \\
			\textit{w.r.t.} EventDance~\cite{zheng2024eventdance} & & & \textbf{+0.95} & & \textbf{+0.07} & & \textbf{+0.20} \\
			\midrule
			Baseline &  & 61.35 & - & 43.59 & - & 36.54 & -  \\ 
			SHOT ~\cite{SHOT} &  & 68.46 & +7.11 & 45.03 & +1.44 &36.18 & -0.36  \\ 
			Zhao \etal ~\cite{CTN}&  & 68.02 & +6.67 & 43.84 & +0.25 & 34.56 & -1.98 \\ 
			SHOT++ ~\cite{SHOT++} & MobileNetV2-Small & 68.40 & +7.05 & 44.90 & +1.31 & 36.05 & -0.49 \\ 
			EventDance~\cite{zheng2024eventdance} & & 70.36 & +9.01 & 51.98 & +8.39 & 44.12 & +7.58  \\ 
			\rowcolor{gray!15} EventDance++ (Ours)  & & 80.63 & +19.28 & 76.82 & +33.23 & 76.64 & +4.01   \\ 
			\textit{w.r.t.} EventDance~\cite{zheng2024eventdance} & & & \textbf{+10.27} & & \textbf{+24.84} & & \textbf{+32.52} \\
			\bottomrule
	\end{tabular}}
	\label{tab:MNIST-to-N-MNIST}
\end{table*}
\begin{figure*}[t!]
	\centering
	\includegraphics[width=\textwidth]{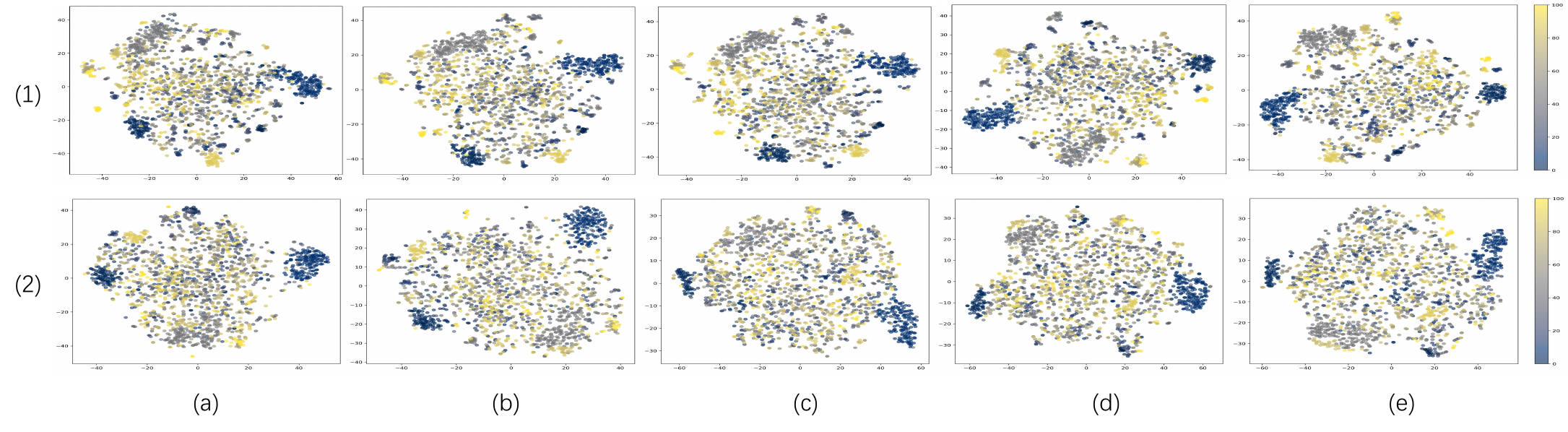}
	\vspace{-16pt}
	\caption{tSNE visualization of different methods on NCaltech-101 dataset: (a) SHOT~\cite{SHOT}, (b) SHOT++~\cite{SHOT++}, (c) Zhao \etal~\cite{CTN}, (d) EventDance~\cite{zheng2024eventdance}, and (e) EventDance++, with (1) Swin-tiny and (2) MobileNetV2-small as backbones.
	}
	\label{fig:tsne_caltech}
	\vspace{-12pt}
\end{figure*}

\section{Experiments}
In this section, we empirically validate various aspects of EventDance++. In Sec.~\ref{sec: Datasets and Implementation Details}, we show the experimental settings of our image-to-event adaptation, baseline, comparison methods, and implementation details. We further show the performance of EventDance++ compared with the existing cross-modal and UDA methods in Sec.~\ref{sec: Experimental Results}.

\begin{table*}[ht!]
	\centering
	\caption{Experimental results on images-to-events (Caltech101-to-N-Caltech101) with \textbf{SFUDA} methods. 
		$\Delta$: The performance gain over the baseline. 
		The \textbf{bold} and \underline{underline} denote the best and the second-best performance in SFUDA methods, respectively.}
	\renewcommand{\tabcolsep}{12pt}
	\resizebox{\linewidth}{!}{
		\begin{tabular}{lcclcccl}
			\toprule
			Method & Backbone & Accuracy & $\Delta$ & Recall & $\Delta$ & F1 & $\Delta$ \\ \midrule
			Baseline & & 46.83 & - & 33.31 & - & 33.90 & -  \\ 
			SHOT ~\cite{SHOT} & & 46.76 & -0.07 & 31.18 & -2.13 & 30.67 & -3.23 \\ 
			Zhao \etal ~\cite{CTN} & & 47.46 & +0.63 & 30.11 & -3.20 & 30.06 & -3.84 \\ 
			SHOT++ ~\cite{SHOT++} & ResNet-18 & 50.53 & +3.70 & 29.49 & -3.82 & 30.37 & -3.53  \\ 
			EventDance~\cite{zheng2024eventdance} & & 66.52 & +19.69 & 35.28 & +1.97 & 35.27 & +1.37 \\ 
			\rowcolor{gray!15} EventDance++ (Ours) & & 71.19 & +24.36 & 38.27 & +4.96 & 40.05 & +6.15 \\ 
			\textit{w.r.t.} EventDance~\cite{zheng2024eventdance} & & & \textbf{+4.67} & & \textbf{+2.99} & & \textbf{+4.78} \\
			\midrule
			Baseline & & 57.94 & - & 34.67 & - & 37.52 & -  \\ 
			SHOT ~\cite{SHOT} &  & 67.98 & +10.40 & 39.79 & +5.12 & 42.57 & +5.05 \\ 
			Zhao \etal ~\cite{CTN}&  & 66.15 & +8.21 & 39.17 & +4.50 & 42.04 & +4.52 \\ 
			SHOT++ ~\cite{SHOT++} & Swin-tiny & 67.35 & +9.41 & 39.56 & +4.89 & 42.25 & +4.73 \\ 
			EventDance~\cite{zheng2024eventdance} & & 68.55 & +10.61 & 40.29 & +5.62 & 43.57 & +6.05  \\ 
			\rowcolor{gray!15} EventDance++ (Ours)  &  & 74.69 & +16.75 & 46.20 & +11.53 & 49.77 & +12.25  \\ 
			\textit{w.r.t.} EventDance~\cite{zheng2024eventdance} & & & \textbf{+6.14} & & \textbf{+5.91} & & \textbf{+6.20} \\
			\midrule
			Baseline & & 65.95 & - & 42.26 & - & 45.95 & - \\
			SHOT ~\cite{SHOT} &  & 70.83 & +4.88 & 43.16 & +0.87 & 45.92 & -0.03  \\ 
			Zhao \etal ~\cite{CTN}&  & 71.22 & +5.27 & 43.89 & +1.63 & 46.83 & +0.88 \\ 
			SHOT++ ~\cite{SHOT++} & ConvNext-tiny & 71.53 & +5.58 & 43.33 & +1.07 & 47.95 & +2.00 \\ 
			EventDance~\cite{zheng2024eventdance} &  & 72.13 & +6.18 & 43.44 & +1.18 & 46.60 & +0.65  \\ 
			\rowcolor{gray!15} EventDance++ (Ours)  &  & 78.55 & +12.60 & 53.24 & +10.98 & 56.63 & +10.68  \\ 
			\textit{w.r.t.} EventDance~\cite{zheng2024eventdance} & & & \textbf{+6.42} & & \textbf{+9.80} & & \textbf{+10.03} \\
			\midrule
			Baseline & & 48.12 & - & 26.79 & - & 28.45 & - \\ 
			SHOT ~\cite{SHOT} &  & 52.44 & +4.32 & 29.21 & +2.42 & 31.16 & +2.71  \\ 
			Zhao \etal ~\cite{CTN}&  & 54.75 & +6.63 & 31.69 & +4.90 & 33.26 & +4.81  \\ 
			SHOT++ ~\cite{SHOT++} & MobileNetV2-Small  & 54.68 & +6.56 & 31.63 & +4.84 & 32.89 & +4.44  \\ 
			EventDance~\cite{zheng2024eventdance} &  & 58.10 & +9.98 & 35.12 & +8.33 & 37.12 & +8.67  \\ 
			\rowcolor{gray!15} EventDance++ (Ours) &  & 63.03 & +14.91 & 36.41 & +9.62 & 39.15 & +10.70  \\ 
			\textit{w.r.t.} EventDance~\cite{zheng2024eventdance} & & & \textbf{+4.93} & & \textbf{+1.29} & & \textbf{+2.03} \\
			\bottomrule
	\end{tabular}}
	\label{tab:Caltech101-to-N-Caltech101}
\end{table*}

\subsection{Datasets and Implementation Details}
\label{sec: Datasets and Implementation Details}
\noindent \textbf{N-MNIST}\cite{dataset}: This event-based rendition of the renowned MNIST dataset is generated by recording an event camera's visual input as it is presented with the original MNIST digits on a monitor. 

\noindent \textbf{N-CALTECH101}\cite{dataset}: Serving as the event-based counterpart to the CALTECH101 dataset, N-CALTECH101 includes 100 object classes and an additional background class. The dataset is particularly challenging due to its diversity of classes and the varying sample sizes within each class.

\noindent \textbf{NCIFAR10}\cite{dataset}:
NCIFAR10 is an event-based version of the CIFAR-10 dataset, which consists of 10 classes representing various objects and animals. The event-based data is captured by recording an event camera as it views the CIFAR-10 images displayed on a screen.

\noindent \textbf{Evaluation.} 
EventDance++ utilizes three distinct target models during the training phase as EventDance~\cite{zheng2024eventdance}. 
The recognition performance of the target models, specifically those processing the voxel grid representation, is presented across three event-based benchmarks in Tab.~\ref{tab:MNIST-to-N-MNIST},~\ref{tab:Caltech101-to-N-Caltech101}, and~\ref{tab:CIFAR10-to-N-CIFAR10}.

\noindent \textbf{Metrics.}
Accuracy is a widely used metric for evaluating classification models. It measures the proportion of correct predictions out of the total number of predictions made. 
Accuracy is a useful metric when the classes are balanced. However, it can be misleading when dealing with imbalanced datasets, as it does not account for the distribution of different classes.
Recall, also known as sensitivity or true positive rate, is a metric that measures the proportion of actual positive instances that are correctly identified by the model. It is particularly useful in scenarios where it is important to minimize false negatives. 
High recall indicates that the model successfully identifies most of the positive instances, making it a critical metric in applications such as medical diagnostics, where missing a positive case can have severe consequences.
The F1 score is a harmonic mean of precision and recall, providing a single metric that balances the trade-off between these two measures. It is particularly useful when dealing with imbalanced datasets, as it considers both false positives and false negatives. 
The F1 score ranges from 0 to 1, with 1 indicating perfect accuracy and recall. It provides a more balanced evaluation of model performance, especially in cases where accuracy and recall are important, and a balance between the two is desired.

\noindent \textbf{Baseline and comparison methods.}
As we are the first to address the cross-modal problem, there is no direct baseline available for comparison. We establish a baseline in all tables to evaluate the performance of the pre-trained source model with event voxel grids, as in the previous work~\cite{CTN}. Also, we compare our method with the existing SFUDA methods~\cite{SHOT, SHOT++}, image-to-voxel-grid adaptation method~\cite{CTN}, and the UDA method using source data~\cite{DSAN} that use event voxel grids as the target modality data. 
\\
\noindent \textbf{Implementation Details.}
In EventDance++, we extend the capabilities of EventDance by incorporating more up-to-date backbones from both transformer and CNN models. Specifically, we utilize ResNet-18~\cite{he2016deep}, EfficientNet~\cite{tan2019efficientnet}, ConvNeXt-tiny~\cite{liu2022convnet}, MobileNetV2-Small~\cite{sandler2018mobilenetv2}, and Swin-transformer~\cite{liu2021swin}. This expansion aims to investigate how the backbone models influence the source-free cross-modal knowledge adaptation efficiency.
We set the batch size to 512 and re-train all the comparison methods originally evaluated in EventDance. For optimization, we employ the AdamW optimizer with an initial learning rate of \(5 \times 10^{-5}\), which linearly decays over time. This optimization strategy ensures stable and efficient convergence.

To enhance the robustness of the source modality pre-training, we apply image augmentation techniques such as random rotations and flipping. These augmentations help in improving the model's generalization capabilities. However, to maintain fairness in comparison with other methods, we refrain from using event augmentation techniques during the target learning phase.
By integrating these improvements, our implementation aims to provide a more comprehensive and rigorous evaluation framework, leveraging state-of-the-art deep learning models.

\begin{figure*}[t!]
	\centering
	\includegraphics[width=\textwidth]{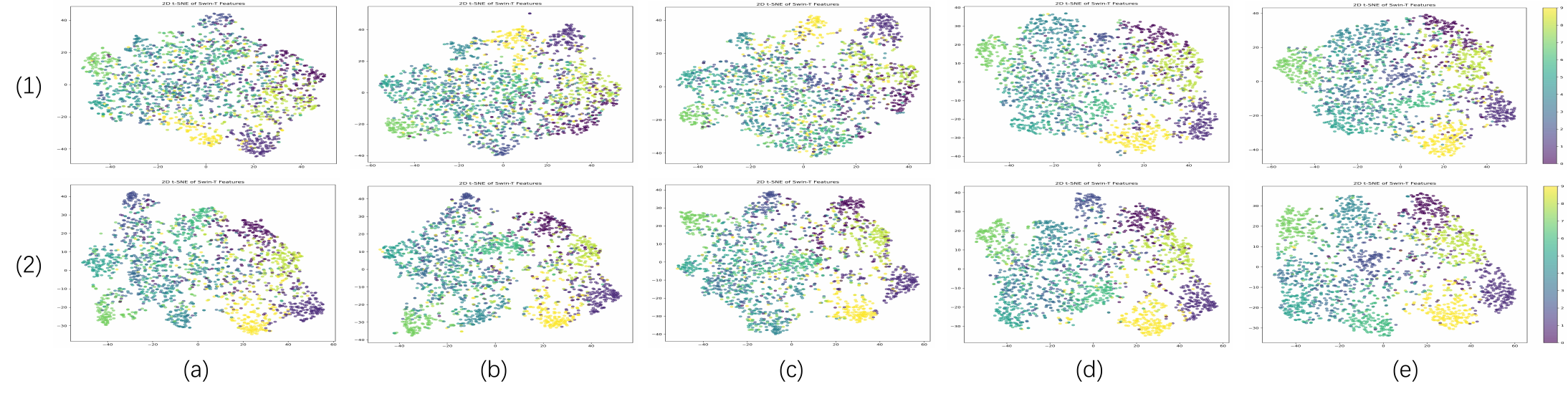}
	\vspace{-16pt}
	\caption{tSNE visualization of different methods on NCIFAR10 dataset: (a) SHOT~\cite{SHOT}, (b) SHOT++~\cite{SHOT++}, (c) Zhao \etal~\cite{CTN}, (d) EventDance~\cite{zheng2024eventdance}, and (e) EventDance++, with (1) Swin-tiny and (2) ConvNeXt-tiny as backbones.
	}
	\label{fig:tsne_compare_cifar}
	\vspace{-12pt}
\end{figure*}

\begin{table*}[ht!]
	\centering
	\caption{Experimental results on images-to-events (CIFAR10-to-N-CIFAR10) with \textbf{SFUDA} methods. 
		$\Delta$: The performance gain over the baseline. 
		The \textbf{bold} and \underline{underline} denote the best and the second-best performance in SFUDA methods, respectively.}
	\renewcommand{\tabcolsep}{12pt}
	\resizebox{\linewidth}{!}{
		\begin{tabular}{lcclcccl}
			\toprule
			Method & Backbone & Accuracy & $\Delta$ & Recall & $\Delta$ & F1 & $\Delta$ \\ \midrule
			Baseline & & 56.26 & - & 47.90 & - & 46.82 & -  \\ 
			SHOT ~\cite{SHOT} & & 62.90 & +6.64 & 45.35 & -2.55 & 44.14 & -2.68 \\ 
			Zhao \etal ~\cite{CTN} & & 62.86 & +6.60 & 46.45 & -1.45 & 45.71 & -1.11 \\ 
			SHOT++ ~\cite{SHOT++} & ResNet-18 & 63.88 & +7.62 & 46.00 & -1.90 & 45.08 & -1.74 \\ 
			EventDance~\cite{zheng2024eventdance} & & 70.47 & +14.21 & 59.65 & +11.75 & 60.31 & +13.49 \\ 
			\rowcolor{gray!15} EventDance++ (Ours)  & & 72.53 & +16.27 & 64.65 & +16.75 & 65.49 & +18.67 \\ 
			\textit{w.r.t.} EventDance~\cite{zheng2024eventdance} & & & \textbf{+2.06} & & \textbf{+5.00} & & \textbf{+5.18} \\
			\midrule
			Baseline & & 53.71 & - & 39.45 & - & 40.04 & -  \\ 
			SHOT ~\cite{SHOT} & & 62.00 & +8.29 & 42.75 & +3.30 & 43.89 & +3.85 \\ 
			Zhao \etal ~\cite{CTN}& & 61.77 & +8.06 & 42.60 & +3.15 & 44.06 & +4.02  \\ 
			SHOT++ ~\cite{SHOT++} & Swin-tiny & 61.88 & +8.17 & 42.50 & +3.05 & 44.01 & +3.97 \\ 
			EventDance~\cite{zheng2024eventdance} & & 70.34 & +16.63 & 58.30 & +18.85 & 60.23 & +20.19 \\ 
			\rowcolor{gray!15} EventDance++ (Ours)  & & 74.20 & +20.49 & 67.40 & +27.95 & 68.59 & +28.55 \\ 
			\textit{w.r.t.} EventDance~\cite{zheng2024eventdance} & & & \textbf{+3.86} & & \textbf{+9.10} & & \textbf{+8.36} \\
			\midrule
			baseline & & 58.16 & - & 48.85 & - & 48.67 & - \\
			SHOT ~\cite{SHOT} & & 66.75 & +8.59 & 53.80 & +4.95 & 54.75 & +6.08  \\ 
			Zhao \etal ~\cite{CTN}& & 68.47 & +10.31 & 53.80 & +4.95 & 54.46 & +5.79  \\ 
			SHOT++ ~\cite{SHOT++} & ConvNext-tiny & 67.22 & +9.06 & 53.95 & +5.10 & 54.50 & +5.83  \\ 
			EventDance~\cite{zheng2024eventdance} & & 70.47 & +12.31 & 59.65 & +10.80 & 60.31 & +11.64   \\ 
			\rowcolor{gray!15} EventDance++ (Ours) & & 72.53 & +14.37 & 64.65 & +15.80 & 65.49 & +16.82  \\ 
			\textit{w.r.t.} EventDance~\cite{zheng2024eventdance} & & & \textbf{+2.06} & & \textbf{+5.00} & & \textbf{+5.18} \\
			\midrule
			Baseline &  & 41.80 & - & 27.30 & - & 26.51 & -  \\
			SHOT ~\cite{SHOT} & & 59.04 & +17.24 & 12.55 & -14.75 & 6.77 & -19.74   \\ 
			Zhao \etal ~\cite{CTN}&  & 59.65 & +17.85 & 13.05 & -14.25 & 7.36 & -19.15 \\ 
			SHOT++ ~\cite{SHOT++} & MobileNetV2-Small & 59.58 & +17.78 & 13.05 & -14.25 & 7.36 & -19.15 \\ 
			EventDance~\cite{zheng2024eventdance} & & 67.95 & +26.15 & 19.70 & -7.60 & 17.94 & -8.57  \\ 
			\rowcolor{gray!15} EventDance++ (Ours)  & & 69.45 & +27.65 & 58.80 & +31.50 & 60.32 & +33.81   \\ 
			\textit{w.r.t.} EventDance~\cite{zheng2024eventdance} & & & \textbf{+1.50} & & \textbf{+39.10} & & \textbf{+42.38} \\
			\bottomrule
	\end{tabular}}
	\label{tab:CIFAR10-to-N-CIFAR10}
\end{table*}

\subsection{Experimental Results}
\label{sec: Experimental Results}

We evaluate EventDance under the challenging source-free image-to-events adaptation setting. The experimental results are shown in Tab.~\ref{tab:MNIST-to-N-MNIST}. EventDance consistently outperforms source-free UDA methods~\cite{SHOT, SHOT++}, the source-free cross-modal UDA method~\cite{CTN}, and achieves recognition accuracy close to the UDA method DSAN~\cite{DSAN} that utilizes source data on the MNIST-to-N-MNIST benchmark.

EventDance++ demonstrates substantial performance improvements across various backbones. Using the ResNet-18 backbone, EventDance++ achieves 75.33\% accuracy, 51.72\% recall, and 54.14\% F1 score, representing gains of +34.30\%, +24.67\%, and +27.11\% over the baseline. Compared to EventDance~\cite{zheng2024eventdance}, EventDance++ improves by +4.98\%, +5.63\%, and +13.59\% in accuracy, recall, and F1 score. With the EfficientNet backbone, EventDance++ achieves 85.16\% accuracy, 83.73\% recall, and 83.26\% F1 score, showing gains of +19.30\%, +23.89\%, and +26.55\% over the baseline, and additional improvements over EventDance of +0.22\%, +0.27\%, and +0.28\%. For the ConvNext-tiny backbone, EventDance++ attains 71.31\% accuracy, 52.05\% recall, and 44.32\% F1 score, with gains of +13.95\%, +9.36\%, and +9.37\% over the baseline, and improvements over EventDance of +0.95\%, +0.07\%, and +0.20\%. Utilizing the MobileNetV2-Small backbone, EventDance++ achieves 80.63\% accuracy, 76.82\% recall, and 76.64\% F1 score, representing gains of +19.28\%, +33.23\%, and +40.01\% over the baseline, and improvements over EventDance of +10.27\%, +24.84\%, and +32.52\%.

As shown in Tab.~\ref{tab:Caltech101-to-N-Caltech101}, EventDance++ achieves substantial improvements across various backbones on the Caltech101-to-N-Caltech101 benchmark with 101 semantic classes. For instance, using the Swin-tiny backbone, EventDance++ achieves 74.69\% accuracy, 46.20\% recall, and 49.77\% F1 score, with gains of +16.75\%, +11.53\%, and +12.25\% over the baseline, and additional gains over EventDance of +16.75\%, +11.53\%, and +12.25\%. Fig.~\ref{fig:tsne_caltech} provides a tSNE visualization of feature distributions obtained by different methods on the NCIFAR10 dataset. EventDance++ demonstrates superior feature clustering, highlighting its enhanced capability in learning discriminative features in the high-level representation space.

Tab.~\ref{tab:CIFAR10-to-N-CIFAR10} further demonstrates significant performance gains of EventDance++. Using the ResNet-18 backbone, EventDance++ achieves 72.53\% accuracy, 64.65\% recall, and 65.49\% F1 score, with gains of +16.27\%, +16.75\%, and +18.67\% over the baseline. With the Swin-tiny backbone, EventDance++ achieves 74.20\% accuracy, 67.40\% recall, and 68.59\% F1 score, showing gains of +20.49\%, +27.95\%, and +28.55\% over the baseline, and improvements over EventDance of +3.86\%, +9.10\%, and +8.36\%. Fig.~\ref{fig:tsne_compare_cifar} provides a tSNE visualization of feature distributions obtained by different methods on the NCIFAR10 dataset. EventDance++ with both Swin-tiny and ConvNext-tiny backbones achieves better feature clustering, indicating superior ability in learning discriminative features for the source-free cross-modal task.

\begin{table}[t!]
\centering
\caption{Experimental results of EventDance++ compared with methods using CLIP.}
\renewcommand{\tabcolsep}{6pt}
\resizebox{\linewidth}{!}{
\begin{tabular}{lccccc}
\toprule 
Method & S.F. & Unsup. & Backbone / Train & N-CAL \\ \midrule
E2VID~\cite{E2VID} & \XSolidBrush & \Checkmark & Fine-tune & 59.80 \\ 
 + CLIP& \XSolidBrush & \Checkmark & Scratch & 9.40 \\ \midrule
Ev-LaFOR~\cite{cho2023label} & \XSolidBrush & \Checkmark  & Text Prompt & 82.46  \\ 
 + CLIP& \XSolidBrush & \Checkmark & Visual Prompt & 82.61 \\ 
\midrule
\multirow{3}{*}{Wang \etal~\cite{wang2019event}} & \Checkmark & \Checkmark & - & 42.70 \\
& \XSolidBrush & \Checkmark & - & 43.50  \\ 
 + CLIP & \Checkmark & \XSolidBrush & - & 39.70 \\ \midrule
& \Checkmark & \Checkmark & R-18 & 71.19  \\ 
EventDance++ & \Checkmark & \Checkmark & Swin-tiny & 74.69 \\ 
& \Checkmark & \Checkmark & ConvNeXt-tiny & 78.55 \\ 
\bottomrule
\end{tabular}}
\label{tab:labelfree}
\end{table}

As shown in Tab.~\ref{tab:labelfree}, we compare our EventDance++ with various label-free methods and methods using CLIP, including E2VID~\cite{E2VID} + CLIP, Ev-LaFOR~\cite{cho2023label} + CLIP, and Wang et al.~\cite{wang2019event} + CLIP. EventDance++ achieves comparable performance to these label-free methods using CLIP, utilizing much lighter neural networks, and demonstrates superior performance without accessing the source modality data.

\begin{table}[t!]
\centering
\caption{Experimental results of EventDance++ compared with supervised methods.}
\renewcommand{\tabcolsep}{6pt}
\resizebox{\linewidth}{!}{
\begin{tabular}{lcccc}
\toprule 
Method & S.F. & Unsup. & Backbone & N-MNIST \\ \midrule
EV-VGCNN~\cite{deng2022voxel} & \XSolidBrush & \XSolidBrush & EV-VGCNN & 99.10   \\ 
Deep SNN~\cite{lee2016training} & \XSolidBrush & \XSolidBrush & Deep SNN & 98.70 \\
Phased LSTM~\cite{neil2016phased} & \XSolidBrush & \XSolidBrush & Phased LSTM & 97.30 \\
PointNet++~\cite{wang2019space} & \XSolidBrush & \XSolidBrush & PointNet++ & 95.50 \\
EventDance & \Checkmark & \Checkmark & R-18 & 71.00  \\ \midrule
& \Checkmark & \Checkmark & EfficientNet & 85.16  \\ 
EventDance++ & \Checkmark & \Checkmark & R-18 & 75.33 \\ 
& \Checkmark & \Checkmark & ConNeXt-tiny & 71.31 \\ 
\bottomrule
\end{tabular}}
\label{tab:supervised}
\end{table}

Furthermore, we compare the performance of our EventDance++ with several state-of-the-art supervised event-based recognition methods on the N-MNIST dataset, as shown in Tab.~\ref{tab:supervised}, including EV-VGCNN~\cite{deng2022voxel}, Deep SNN~\cite{lee2016training}, Phased LSTM~\cite{neil2016phased}, and PointNet++~\cite{wang2019space}. Our EventDance++ achieves competitive performance in an unsupervised manner, without using the source data.

\begin{table*}[t!]
\centering
\caption{Ablation study of different module combinations on N-CALTECH101 with ResNet-18.}
\setlength{\tabcolsep}{10pt}
\resizebox{\linewidth}{!}{
\begin{tabular}{cccccccccccc}
\toprule
$\mathcal{L}_{Sup}$ & $\mathcal{L}_{EN}$ & $\mathcal{L}_{TC}$ & $\mathcal{L}_{PC}$ & $\mathcal{L}_{R}$ & $\mathcal{L}_{KD}$ & Acc &$\Delta$  & Recall & $\Delta$ & F1 & $\Delta$ \\ \midrule %
\Checkmark &  &  & &  &  & 47.10 & - & 25.38 & - & 26.54 & - \\ %
\Checkmark & \Checkmark &  & & &  & 50.14 & +3.04 & 25.89 & +0.51 & 27.00 & +0.46  \\ %
\Checkmark &  & \Checkmark & &  &  & 55.47 & +5.33  & 26.17 & +0.79 & 27.49 & +0.95  \\ %
\Checkmark &  &  & \Checkmark &  &  & 56.57 & +9.47  & 26.84 & +1.46 & 28.01 & +1.47  \\ %
\Checkmark &  &  &  & \Checkmark &  & 56.19 & +9.09 & 25.44 & +0.06 & 26.89 & +0.35  \\ 
\Checkmark &  &  &  &  & \Checkmark & 62.65 & +15.55 & 37.50 & +12.12 & 39.47 & +12.93  \\ 
\Checkmark & \Checkmark & \Checkmark & \Checkmark & \Checkmark & \Checkmark & 71.19 & +24.09 & 38.27 & +12.89 & 40.05 & +13.51  \\
 \bottomrule
\end{tabular}}
\label{tab:modulecombin}
\end{table*}

\begin{figure*}[t!]
    \centering
    \includegraphics[width=\linewidth]{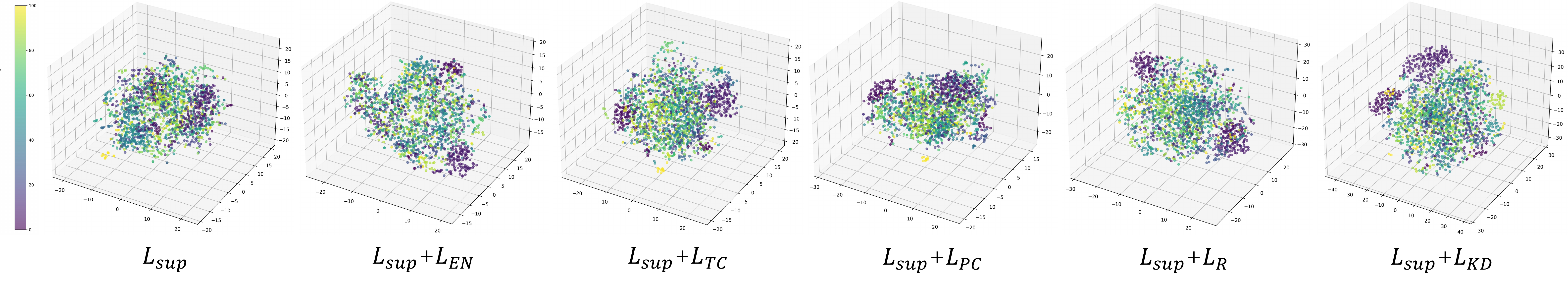}
    \vspace{-16pt}   
    \caption{TSNE~\cite{van2008visualizing} visualization of different loss combinations on N-Caltech101 with ResNet-18, different colors stand for different classes in N-Caltech101.}
    \vspace{-12pt}   
    \label{fig:tsne}
\end{figure*}

\section{Ablation Study and Analysis}
\label{sec: Ablation Study and Analysis}
\subsection{Different Loss Function Combinations}
To validate the effectiveness of the proposed modules, we conducted experiments on the Caltech-to-NCaltech101 benchmark with various combinations of loss functions. Tab.~\ref{tab:modulecombin} presents the performance results with different loss function and component combinations. All proposed modules and loss functions positively impact recognition accuracy. Notably, the combination of all modules, including $L_{sup}$, $L_{EN}$, $L_{TC}$, $L_{PC}$, $L_{R}$, and $L_{KD}$, yields the highest accuracy of 71.19\%, with an improvement of +24.09\% over the baseline. This substantial performance gain supports our claim that incorporating these modules enhances knowledge transfer and recognition accuracy. The t-SNE visualization in Fig.~\ref{fig:tsne} further illustrates the effectiveness of each loss function combination, demonstrating improved feature clustering in the high-level feature space.

\begin{table}[t!]
\centering
\caption{Ablation experiments on the inference of our proposed method with different event representations.}
\setlength{\tabcolsep}{10pt}
\resizebox{\linewidth}{!}{
\begin{tabular}{cccc}
\toprule
\multirow{2}{*}{Backbone} & \multicolumn{3}{c}{Event Representations} \\ \cmidrule{2-4} 
 & Stack Image & Voxel Grid & Event Spike Tensor \\ \midrule
R-18 & $66.70_{\textcolor{blue}{-0.07}}$ & 66.77 & $66.96_{\textcolor{red}{+0.19}}$ \\ \midrule
R-34 & $71.16_{\textcolor{blue}{-1.52}}$ & 72.68 & $73.00_{\textcolor{red}{+0.52}}$ \\ \midrule
R-50 & $91.54_{\textcolor{blue}{-0.81}}$ & 92.35 & $92.74_{\textcolor{red}{+0.39}}$  \\ \bottomrule
\end{tabular}}
\label{Tab: TargetRep}
\end{table}

\subsection{Ablation of Language-guided Reconstruction-based Modality Bridging}
The L-RMB module plays a crucial role in reducing the modality gap between images and events. As shown in Tab.\ref{tab:modulecombin}, a check mark indicates the application of the fine-tuning loss $L_R$. When utilizing only the pre-trained reconstruction model $F_R$ to generate surrogate data, the recognition accuracy reaches 47.10\%. However, by updating $F_R$ through optimization with $L_R$, the accuracy improves substantially to 56.19\%. Additionally, the qualitative results presented in Fig.~\ref{fig:Recon_com} demonstrate that the quality of the reconstructed images progressively improves during the knowledge adaptation process.


\begin{table}[t!]
\centering
\caption{Ablation study of different module combinations on N-CALTECH101 with ResNet-18.}
\setlength{\tabcolsep}{2pt}
\resizebox{\linewidth}{!}{
\begin{tabular}{ccccccccc}
\toprule
$\mathcal{L}_{Sup}$ & $\mathcal{L}_{VKD}$ & $\mathcal{L}_{PKD}$ & Acc &$\Delta$  & Recall & $\Delta$ & F1 & $\Delta$ \\ \midrule %
\Checkmark &  &  & 47.10 & - & 25.38 & - & 26.54 & - \\ %
\Checkmark & \Checkmark &  & 62.42 & +15.32 & 32.41 & +7.03 & 33.24 & +6.70  \\ %
\Checkmark &  & \Checkmark & 61.66 & +14.56 & 32.47 & +7.09 & 33.42 & +6.88 \\ %
\Checkmark & \Checkmark & \Checkmark & 62.65 & +15.55 & 37.50 & +12.12 & 39.47 & +12.93 \\ \bottomrule
\end{tabular}}
\label{tab: KD}
\end{table}

\noindent \textbf{Ablation of $\mathcal{L}_{KD}$.} Considering that $\mathcal{L}_{KD}$ encompasses dual-level knowledge distillation through $\mathcal{L}_{VKD}$ and $\mathcal{L}_{PKD}$, we conduct ablation experiments on the N-Caltech101 dataset using the ResNet-18 model to evaluate the effectiveness of these components. The objective is to systematically assess the individual contributions of $\mathcal{L}_{VKD}$ and $\mathcal{L}_{PKD}$ to the overall performance. Quantitative results are presented in Tab.~\ref{tab: KD}. The incremental addition of $\mathcal{L}_{VKD}$ and $\mathcal{L}_{PKD}$ to the supervisory loss $\mathcal{L}_{Sup}$ yields significant gains in accuracy, recall, and F1 score over the baseline. Notably, the combined use of $\mathcal{L}_{VKD}$ and $\mathcal{L}_{PKD}$ results in the highest performance boost, suggesting that the dual-level knowledge distillation approach effectively leverages the complementary strengths of both components.

\subsection{Ablation of Multi-representation Knowledge Adaptation}
\noindent\textbf{Event representation vs. target model's performance.}
For a fair comparison, we validate the quantitative results of all methods using event voxel grids. To investigate how to fully leverage the abundant spatio-temporal information of events for object recognition, we provide the results of validating EventDance with different representation types in Tab.~\ref{Tab: TargetRep}. 
Compared to inference with voxel grids, using EST, which contains more spatio-temporal information of events, achieves the best recognition accuracy gains by \textbf{+0.19\%, +0.52\%, and +0.39\%}, on the backbones R-18, R-34, and R-50, respectively. The results reflect that EST is better suited for object recognition.
The stack images, for which temporal information is lost, achieve lower recognition accuracy than the voxel grids by \textbf{-0.07\%, -1.52\%, and -0.81\%} with backbones R-18, R-34, and R-50, respectively. Thus, it is crucial to explore the event data with representations that remain temporal characteristic of events in the cross-modal adaptation problems.

\begin{table}[t!]
\centering
\caption{Ablation on the usage of event representations in target model training. (S: stack image; V: voxel grid; E: EST.)}
\setlength{\tabcolsep}{12pt}
\resizebox{\linewidth}{!}{
\begin{tabular}{ccccc}
\toprule
\begin{tabular}[c]{@{}c@{}}Representation\end{tabular} & S & V & E & All \\ \midrule
Accuracy & 61.63 & 63.58 & 65.74 & \textbf{66.91} \\ \bottomrule
\end{tabular}}
\label{Tab:repcomnbin}
\end{table}

\noindent \textbf{Event representation vs. prediction consistency.}
We investigate how different event representation types impact the effectiveness of prediction consistency of the KA module. We test different combinations, including using a single event representation type and multiple representation types. As shown in Tab.~\ref{Tab:repcomnbin}, using all three event representations (stack images, voxel grid, EST) together achieves the best recognition accuracy of \textbf{66.91\%}. The reason is that using three event representations not only enables learning more spatio-temporal information from events but also imposes effective data augmentation for consistency regularization.  

\noindent \textbf{Ablation of cross-modal learning.}
Most of the existing source-free cross-modal adaptation approaches~\cite{CTN, SOCKET} sorely focus on learning the target models while keeping the source model fixed. However, as we discussed in Sec.~\ref{sec:Problem Setup}, 
the source model is not ideal for the image-to-events adaptation problem. Consequently, we introduce the cross-modal learning strategy to simultaneously update source and target models. As shown in Tab.~\ref{tab:modulecombin}, introducing loss functions $\mathcal{L}_{TC}$ and $\mathcal{L}_{KD}$ results in an accuracy increase from 47.10\% to \textbf{55.47\%} -- a significant \textbf{8.37\%} performance gain.

\section{Discussion}
\label{Discussion}
\noindent \textbf{Comparison with ~\cite{messikommer2022bridging}.} 
The approach in ~\cite{messikommer2022bridging} leverages unpaired image and event data for UDA by generating pseudo-optical flow maps from both modalities. These maps are integrated with the image to produce clean events through an event generation model. However, when source domain images are unavailable, the method in ~\cite{messikommer2022bridging} fails to perform image-to-event UDA due to the unreliable event generation process, creating a significant obstacle in training the event encoder for cross-modal tasks. \textit{To overcome this limitation, we introduce the Language-guided Reconstruction-based Modality Bridging (L-RMB) module in EventDance++, which facilitates self-supervised learning to bridge the source and target modalities.}


\noindent \textbf{High Temporal Resolution of Events.}
Cross-modal knowledge transfer is challenging due to the distinct differences between image and event modalities, with images represented as $H \!\times\! W \!\times\! C$ and events as $(x,y,t,p)$~\cite{deeplearningeventsurvey}. A common approach to address this gap is to convert events into image-like tensors. However, many event representations suffer from information loss, particularly temporal information. While certain event representations, such as EST~\cite{EST}, achieve state-of-the-art performance in specific tasks like object recognition, we find that multiple event representations are effective for source-free cross-modal adaptation. This finding is corroborated by the quantitative results shown in Tab.~\ref{Tab:repcomnbin}.

\noindent \textbf{Surrogate data in training.}
While building surrogate data incurs higher computation costs, it effectively eliminates the need for extra paired data that may not always be available in practice. Moreover, the reconstruction model used to construct the surrogate data is only trained and updated during the training phase and can be freely discarded during the inference.

\noindent \textbf{Selection of Anchor Data.}
Through experimentation, we determined that the most effective method for selecting anchor data is to use the first surrogate image reconstructed from the initial period of the event stream. This approach ensures reliable anchor data, especially given the varying lengths of event streams across the target dataset. Subsequent frames are utilized as augmentations for the anchor data. We also explored randomly selecting anchor data, but this resulted in low-quality surrogate images for shorter event streams.

\section{Conclusion}
In this paper, we tackled the novel challenge of image-to-event adaptation for event-based recognition without access to source images. To address this, we introduced EventDance++, a cross-modal framework that integrates the L-RMB and MKA modules for efficient cross-modal knowledge extraction and transfer. Extensive experiments on three benchmark datasets validate the effectiveness of the proposed modules, particularly highlighting the knowledge extraction and modality bridging capabilities of L-RMB, and the spatiotemporal knowledge transfer facilitated by MKA. Our approach successfully bridges the modality gap between images and events, enabling robust event-based object recognition.

\noindent \textbf{Future Work:} 
While EventDance demonstrates promise, training three target models with different representations increases computational costs during training. Despite this, our method has significant implications for event-based vision and could pave the way for new research directions. In the future, we aim to extend our approach to other downstream tasks, improving its practicality and efficiency. We also recognize that while EventDance's performance is comparable to methods utilizing source data, further enhancements are needed to make it suitable for real-world applications. Addressing these challenges will be a key focus of our future work.

{
    \small
    \bibliographystyle{ieeetr}
    \bibliography{main}
}

\begin{IEEEbiography}[{\includegraphics[width=0.9in,height=1.2in,clip,keepaspectratio]{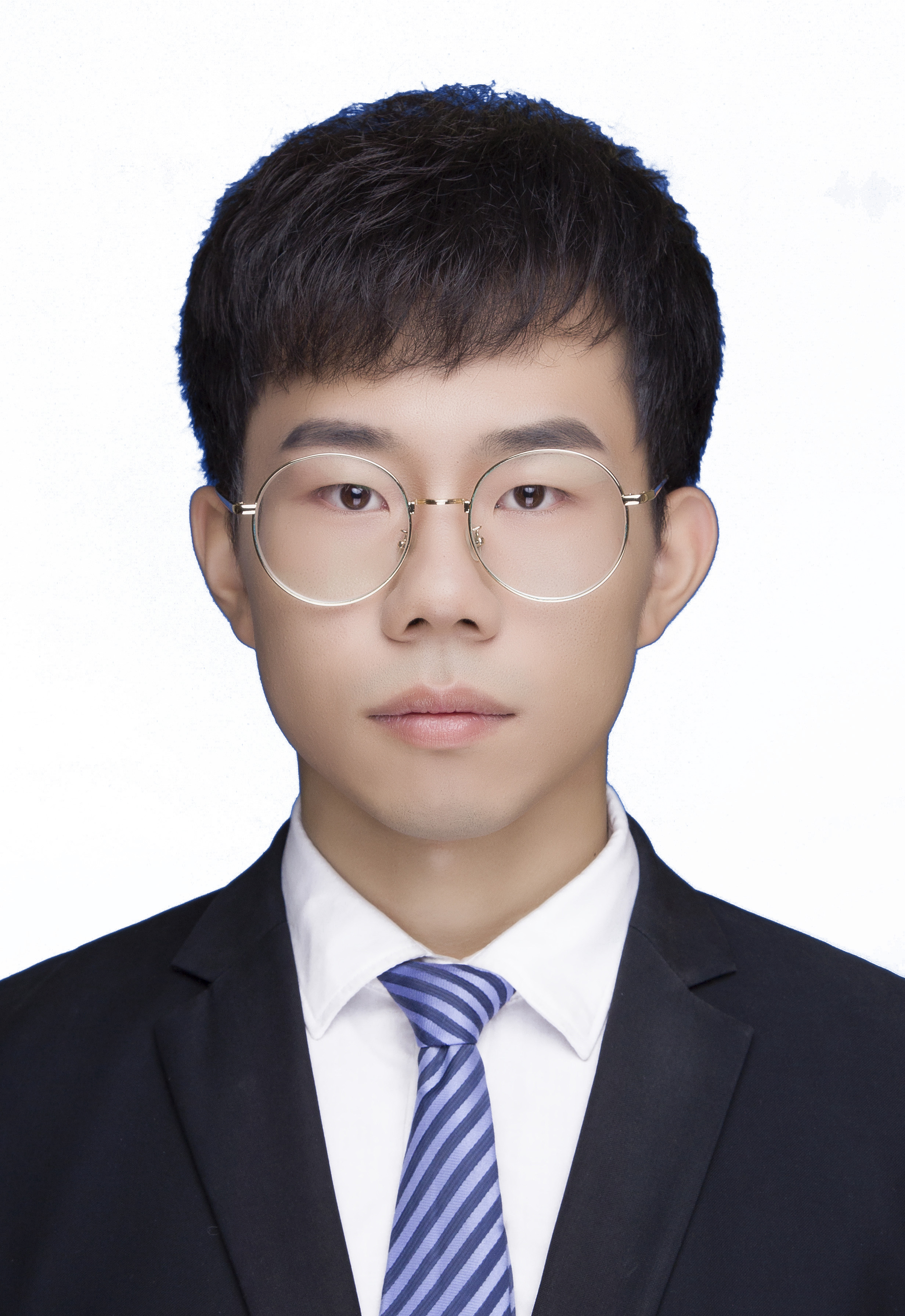}}] {Xu Zheng} (IEEE Student Member) is a Ph.D. student in the Visual Learning and Intelligent Systems Lab,  Artificial Intelligence Thrust, The Hong Kong University of Science and Technology, Guangzhou Campus (HKUST-GZ). He got his B.E. and M.S. from Northeastern University, China. His research interests include multi-modal learning, sensing and perception, \etc.
\end{IEEEbiography}
\vspace{-30pt}
\begin{IEEEbiography}[{\includegraphics[width=0.9in,height=1.2in,clip,keepaspectratio]{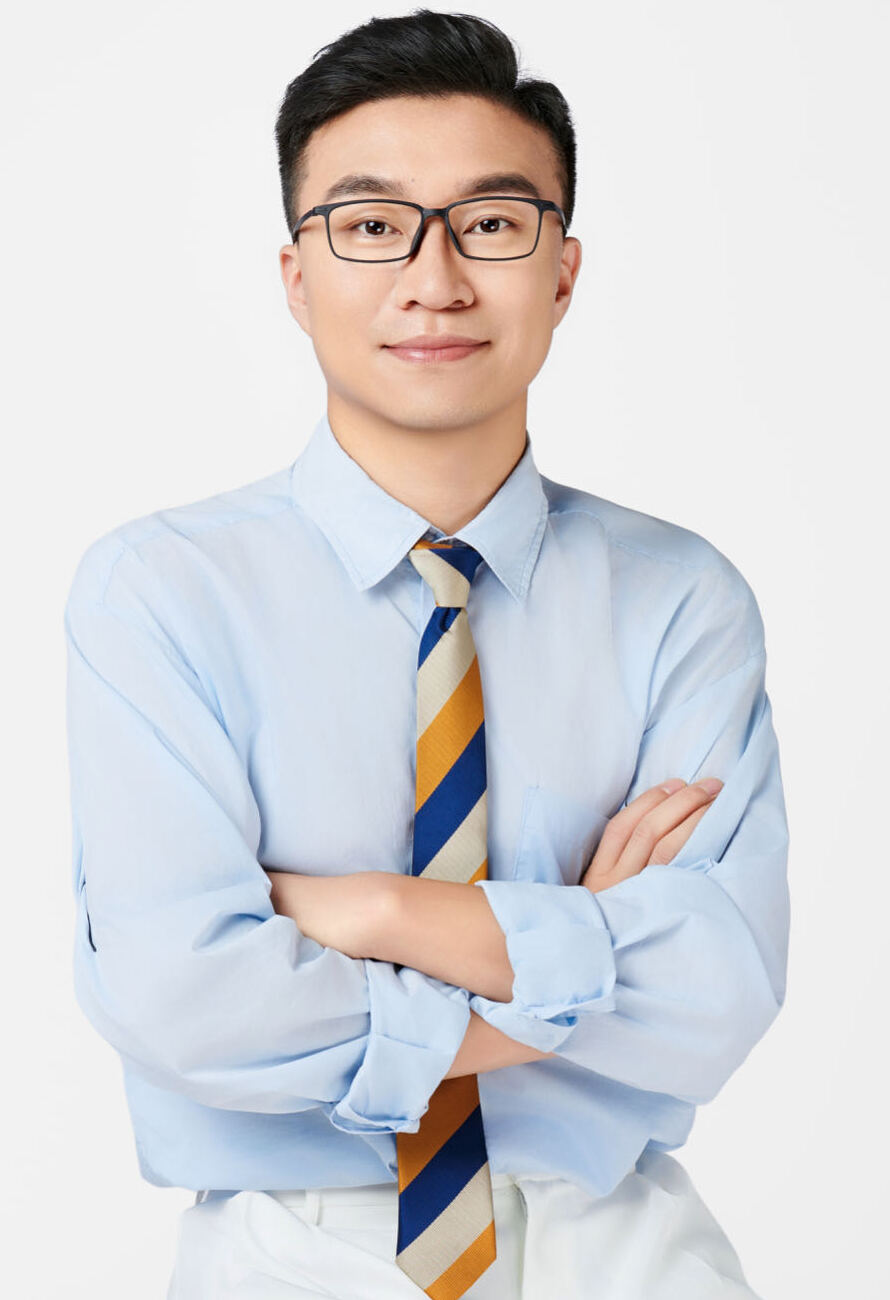}}] 
{Lin Wang} (IEEE Member) is an assistant professor in the AI Thrust, HKUST-GZ, HKUST FYTRI, and an affiliate assistant professor in the Dept. of CSE, HKUST. He did his Postdoc at the Korea Advanced Institute of Science and Technology (KAIST). He got his Ph.D. (with honors) and M.S. from KAIST, Korea. He had rich cross-disciplinary research experience, covering mechanical, industrial, and computer engineering. His research interests lie in computer and robotic vision, machine learning, intelligent systems (XR, vision for HCI), etc. 

\end{IEEEbiography}
\vfill

\end{document}